\documentclass{article}
\usepackage{stywhispers,amsmath,epsfig}

\usepackage{xcolor}
\usepackage{subfigure}
\usepackage{url}
\usepackage{booktabs}
\usepackage{array}
\usepackage{verbatim}
\usepackage[font=footnotesize]{caption}
\usepackage{enumitem}
\usepackage{bbm}
\usepackage{graphicx}
\usepackage{amsmath,amsfonts}
\usepackage{cite}
\usepackage{url}
\usepackage{newtxmath}
\usepackage{booktabs}
\usepackage{algorithm}
\usepackage{algorithmicx}
\usepackage{algpseudocode}
\usepackage{dirtytalk}
\usepackage{empheq}
\usepackage{xurl}
\usepackage{hyperref}

\newcommand{\myitem}[1]{\vspace{0.05\baselineskip}\noindent\textbf{#1}\ }

\graphicspath{{plots/}}
\begin{document}


\title{Minerals in the Wild: A Hyperspectral--XRF Dataset \\ for Elemental Composition Estimation}


\name{Eleftheria Tetoula-Tsonga$^{1}$, George Arvanitakis$^{2}$, Theodoros Giannakas$^{1}$\thanks{This work has received funding from the European Union under Grant Agreement No.~101178897 (DEXPLORE).}}
\address{$^{1}$Institute of Communication and Computer Systems, Athens, Greece\\
$^{2}$Geonova, Athens, Greece}

\maketitle

\begin{abstract}
Rapid mineral characterization is essential for applications ranging from mineral exploration to industrial ore processing.
To this end, Hyperspectral Imaging (HSI) has emerged as a promising sensing modality thanks to its fine spectral resolution, enabling mineral discrimination in both close-range and remote sensing settings. However, the scarcity of publicly available datasets with reliable ground-truth labels hinders the development and evaluation of HSI-based mineral identification methods.
We release \emph{Minerals in the Wild}, a multi-purpose dataset comprising 1,132 rock specimens collected across Europe~\cite{minerals-in-the-wild}. For each specimen, we provide an HSI acquisition together with an elemental characterization obtained via an XRF sensor. 
We define the task of elemental characterization on our dataset and propose a pruning mechanism that removes distant signatures from the USGS dictionary prior to a convex optimization approach for matching HSI pixels with USGS spectral signatures.
Finally, we empirically show that our approach outperforms simpler baselines.
\end{abstract}


\section{Introduction}
\label{sec:intro}
\subsection{Motivation}

The access to critical raw materials is becoming a bottleneck for key technologies--from AI and advanced manufacturing to electrification and clean energy. For example, AI infrastructure relies heavily on mineral-intensive semiconductors and data centers; while clean-energy deployment depends on lithium, cobalt, nickel, copper, graphite, and rare earth elements for batteries and grids~\cite{iea2025energyai}. Traditionally, to \emph{explore} the mineral content of a new area, one needed to set up costly field campaigns, often with highly uncertain reward. Beyond exploration, modern mining workflows increasingly require \emph{automated sorting} of raw material, under stringent latency requirements (often in the order of seconds), rendering the use of manual high-precision equipment impractical~\cite{lessard2014oreSorting}.

To address these challenges \emph{at scale}, the community is moving toward efficient, yet approximate, mineralogical inference using image-based sensing. In particular, the imagery employed in this setting is typically of high spectral resolution, commonly referred to as hyperspectral imaging (HSI), due to the fact that many minerals exhibit informative features, e.g., absorption bands, at wavelengths not captured by conventional RGB. Nonetheless, importantly, the camera on its own is by no means enough; it has to be paired with an identification algorithm capable of inferring the presence of minerals in the image. To devise such algorithms, we typically have to associate two (or potentially more) modalities: (a) hyperspectral image, and (b) matched mineralogical or elemental characterization. While acquiring modality (a) is relatively fast and scalable, acquiring modality (b) requires time-consuming measurements with specialized equipment. Taking the above into account, the importance of such paired datasets emerges naturally, as it enables practitioners and researchers to build a variety of applications related to remote sensing and automated material sorting.

\subsection{Related Work}

\myitem{Datasets.} Hyperspectral mineral datasets span satellite, airborne, and laboratory acquisition regimes. Satellite (e.g., EnMAP) and airborne (e.g., Cuprite/HYDICE) datasets primarily support regional-scale mineral mapping and exploration~\cite{asadzadeh2025leveraging,resmini1997mineral,swayze2014mapping}. Laboratory datasets are closer to our setting; for example, Koirala \emph{et al.} introduced a multisensor benchmark based on controlled mineral powder mixtures~\cite{koirala2023multisensor}. In contrast, our dataset comprises intact natural rock specimens, pairing HSI cubes with XRF-derived elemental compositions to support specimen-level HSI-to-XRF prediction.


\myitem{Unmixing.} 
The problem of pixel unmixing has been thoroughly studied in the context of satellite imaging. Classical approaches include constrained least-squares (sparse) formulations~\cite{heinz2001fully}, with them either decomposing pixels to representative in-image endmembers or using reference spectra (e.g., USGS libraries)~\cite{iordache2011sparse}. More recent approaches incorporate nonlinear modeling and spectral variability~\cite{koirala2020robust,rasti2022hapkecnn}. Our pixel unmixing pipeline builds on~\cite{iordache2011sparse} by introducing a novel library-reduction filter for improved fit during optimization.

\subsection{Contributions and Outline}


\myitem{(C.1)} We release \emph{Minerals in the Wild}, a new multi-purpose dataset of 1,132 solid hand-sized rock specimens collected across Europe~\cite{minerals-in-the-wild}. For each specimen we store: a) HSI, and b) XRF characterization, i.e., the specimen's composition in the form of percentages over chemical elements (Section~\ref{sec:data}).

\myitem{(C.2)} We define a case study on the dataset. In particular, using the USGS library as support data, we estimate the elemental composition of a specimen using its HSI cube. To address the problem, we design an efficient algorithm, based on convex optimization, paired with a novel (regularizing) filter. The algorithm estimates the elemental composition and we validate said estimates against XRF ground truth measurements (Section~\ref{sec:case}).

\section{A Hyperspectral-XRF Dataset}
\label{sec:data}
Physically, \emph{Minerals in the Wild} comprises a collection $\mathcal{S}$ of $S=1,132$ hand-sized specimens gathered from multiple mines and exploration areas across Europe. 
Our first contribution is to convert this collection into a structured dataset. Below, we detail the acquisition protocol for the HSI and the XRF sensing modalities, as well as the data representation of each specimen in the dataset. Finally, we report dataset- and sample-level insights. 
In what follows, tensors of arbitrary size are denoted with bold. For example, $\mathbf{X}\in\mathbb{R}^{A\times B}$ is a 2D matrix, while $\mathbf{X}(i,j)$ is its scalar value at index $(i,j)$.

\subsection{Hyperspectral Modality: Specimens to HSI Cubes} \label{subsec:hsi}

\myitem{Step~(1): Scan.} The setup consists of a conveyor belt and a Specim SWIR hyperspectral camera~\cite{specim_swir}, mounted approximately 30~cm above the belt and oriented towards it. The spatial content of specimen $s$ is acquired as follows. 
At each timestep, the push-broom sensor captures a spatial line of width $W=384$ pixels. As $s$ passes under the sensor, successive lines being accumulated result in an image of size $H_s \times W$, where $H_s$ denotes the number of acquired lines, and depends on its size.
Spectrally, for each pixel the camera stores $B=273$ values, where $B$ is the number of bands, in the 1000--2500~nm wavelength range\footnote{This range is known to contain diagnostic absorption features that facilitate mineral discrimination~\cite{goetz1985imaging}.}.
At the end of Step~(1), the data object stored for specimen $s$ is denoted as $\mathbf{X}^{(1)}_s \in \mathbb{R}^{H_s\times W\times B}$.



\myitem{Step~(2): Calibrate.} Using dark- and white-references, denoted by
$\mathbf{I}_{\text{Dark}},~\mathbf{I}_{\text{White}}\in\mathbb{R}^{W\times B}$, respectively, we calibrate
$\mathbf{X}^{(1)}_s$. These references vary across detector position and wavelength, but remain constant along the scan dimension. Specifically we compute:
\begin{equation}
\mathbf{X}^{(2)}_s(i,j,b)=
\frac{\mathbf{X}^{(1)}_s(i,j,b)-\mathbf{I}_{\text{Dark}}(j,b)}
{\mathbf{I}_{\text{White}}(j,b)-\mathbf{I}_{\text{Dark}}(j,b)},~~~\forall (i,j,b).
\end{equation}

\myitem{Step~(3): Mask-out.} We manually create a binary spatial mask $\mathbf{M}_s\in\{0,1\}^{H_s\times W}$ for specimen $s$, where $\mathbf{M}_s(i,j)=1$ indicates that pixel $(i,j)$ visually belongs to $s$, and $\mathbf{M}_s(i,j)=0$ otherwise. To obtain the final representation, denoted by $\mathbf{X}^{(3)}_s \in \mathbb{R}^{H_s\times W\times B}$, we apply the mask as follows:
\begin{equation}
\mathbf{X}^{(3)}_s(i,j,b)=
\begin{cases}
\mathbf{X}^{(2)}_s(i,j,b), & \text{if } \mathbf{M}_s(i,j)=1,\\
\texttt{NaN}, & \text{if } \mathbf{M}_s(i,j)=0,
\end{cases}
\quad \forall (i,j,b).
\end{equation}
Effectively, $\mathbf{X}^{(3)}_s$ is the HSI cube of specimen $s$ with the background filled with \texttt{NaN} across all $B$ bands. For the rest of this work, we drop the step superscript and $\mathbf{X}_s\equiv\mathbf{X}_s^{(3)}~\forall~s\in\mathcal{S}$.
%
For each specimen $s$, let $\Omega_s$ denote the set of valid pixel locations after masking; we enumerate these by $p = 1,\dots, |\Omega_s|$, and denote the corresponding spectra by $\mathbf{x}_{s,p}$.

\subsection{XRF Modality: Specimens to Compositions} \label{subsec:xrf}

\myitem{Step~(1): Scan.} To characterize specimen $s$, we use X-ray fluorescence (XRF)~\cite{bruker_s1_titan} by performing $D=5$ point measurements at visually different locations of the surface, each with an 8~mm analysis spot. Each measurement returns a vector of $N$ normalized compositions over a mixed set of column labels, including elements (e.g., Mn) and compounds (e.g., MgO). The resulting matrix is denoted by $\mathbf{y}^{(1)}_{s}\in\mathbb{R}^{D\times N}$, where each row corresponds to one location.

\myitem{Step~(2): Average.} To obtain a specimen-level composition, we aggregate by averaging across the $D$ measurements, which yields vector $\mathbf{y}^{(2)}_s\in\mathbb{R}^{N}$.

\myitem{Step~(3): Harmonize.} This step turns the mixed (elements and compounds) label space to element-only space. Compound compositions are decomposed into their elemental contributions using stoichiometric mass fractions\footnote{For example, suppose Mn=0.9 and MgO=0.1. The MgO splits to: Mg as $24.3/(24.3+16.0)$ and O as $16.0/(24.3+16.0)$. These are weighted by 0.1 to yield Mn=0.9, Mg=0.06, and O=0.04. After removing O and renormalizing, the final representation becomes Mn=0.937 and Mg=0.063.}.
The contributions from all sources are then accumulated at element level. The resulting vector is filtered by removing light elements (e.g., O and C) and renormalized to obtain a valid elemental composition (summing to 1). Thus, compound coordinates (e.g., MgO) are replaced by their corresponding elemental coordinates (e.g., Mg), while the vector dimensionality is preserved. The final vector is denoted by $\mathbf{y}^{(3)}_s\in\mathbb{R}^{N}$ ($N=51$ in our case). For the rest of this work, $\mathbf{y}_s\equiv\mathbf{y}^{(3)}_s~\forall~s\in\mathcal{S}$.



\subsection{Discussion and Insights} \label{subsec:analysis}

With the dataset $\mathcal{D}=\{\mathbf{X}_s, \mathbf{y}_s\}_{s\in\mathcal{S}}$ in place, we attempt to export interesting insights, using the two modalities.

\begin{figure}
\centering
\subfigure[Dominant elements]{\includegraphics[width=0.45\columnwidth]{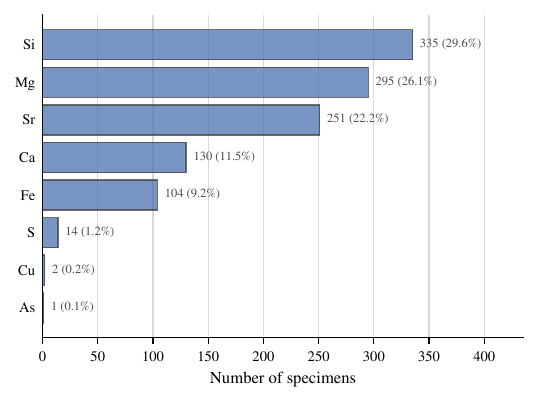}\label{fig:xrf-dominance}}
\hspace{0.08\columnwidth}
\subfigure[Elemental mass fractions]{\includegraphics[width=0.45\columnwidth]{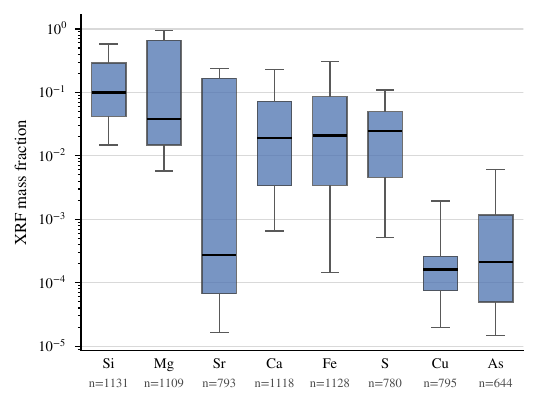}\label{fig:xrf-boxplot}}
\vspace{-2.5mm}
\caption{XRF diversity: (a) Number of specimens for which an element is dominant. (b) Distribution of non-zero elemental mass fractions across specimens.}
\label{fig:diversity}
\end{figure}

\myitem{Elemental diversity (XRF).} To characterize the elemental diversity of $\mathcal{S}$, we focus on our XRF vectors. In Fig.~\ref{fig:xrf-dominance}, we see a histogram showing the number of specimens in which an element dominates. For example, around 30$\%$ of the specimens have Si as their top element. We observe that the absence of a single dominant element suggests that $\mathcal{D}$ spans a broad range of compositions, providing a challenging benchmark for evaluating hyperspectral unmixing methods. In addition, Fig.~\ref{fig:xrf-boxplot} shows a boxplot (in logarithmic scale) of the non-zero observed fractions across the $S$ specimens, strengthening the case that our data exhibit interesting variability when it comes to the observed XRF fractions.

\begin{figure}
\centering
\subfigure[]{\includegraphics[width=0.5\columnwidth]{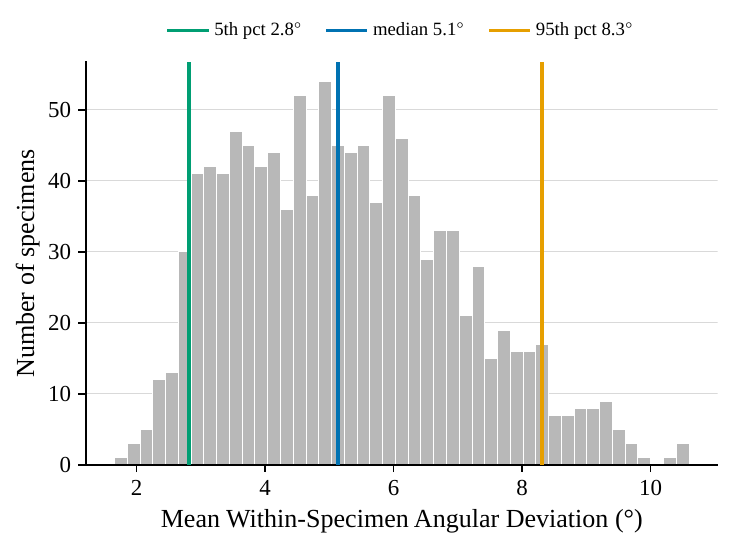}\label{fig:sam-hist}}
\hspace{0.08\columnwidth}
\subfigure[]{\includegraphics[width=0.35\columnwidth]{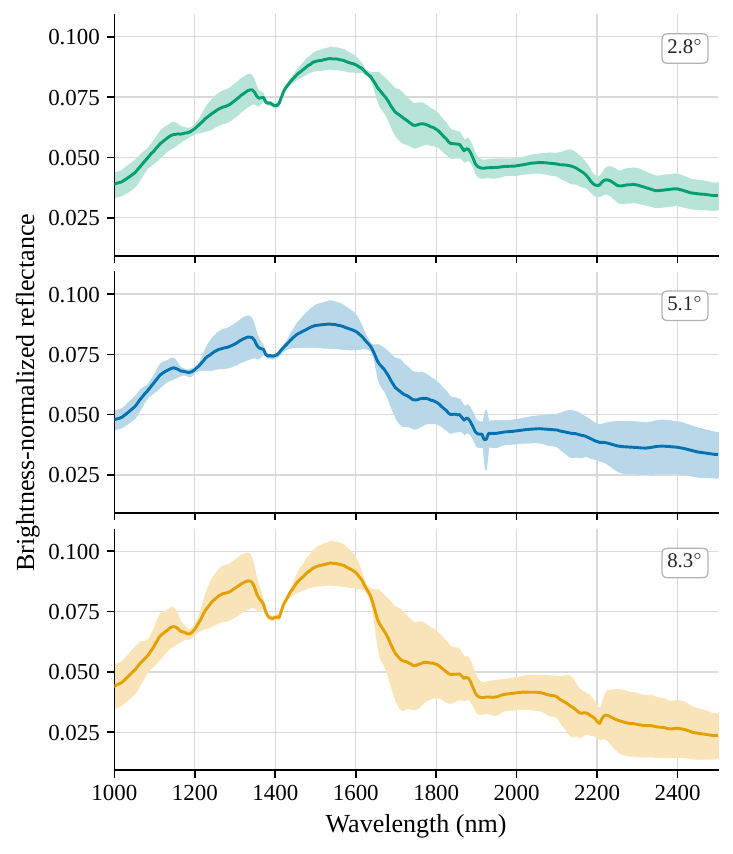}\label{fig:spectral-env}}
\vspace{-2.5mm}
\caption{Pixel-level spectral uniformity: (a) Distribution of the Mean Within-Specimen Angular Deviation (MWAD). (b) Mean spectra with corresponding spectral envelopes for representative specimens.}
\label{fig:uniformity}
\end{figure}

\myitem{Spectral variability (HSI).} Since the camera acquires complete HSI cubes, we examine spectral the variability of $s$ using angular differences between its spectra; 
hence, we define the Mean Within-Specimen Angular Deviation (MWAD) of $s$:
\begingroup
\setlength{\abovedisplayskip}{3pt}
\setlength{\belowdisplayskip}{3pt}
\begin{align}\label{eq:mwad}
\mathrm{MWAD}_s=
\frac{1}{|\Omega_s|}
\sum_{p\in\Omega_s}\theta(\mathbf{x}_{s,p},\bar{\mathbf{x}}_s)
\end{align}
\endgroup
where $\bar{\mathbf{x}}_s$ is the mean spectrum of specimen $s$, and $\theta(\cdot,\cdot)$ denotes the angular deviation between two spectra. Figure~\ref{fig:sam-hist} summarizes its distribution, while Fig.~\ref{fig:spectral-env} shows representative $\bar{\mathbf{x}}_s$ together with their envelopes. Overall, the within-specimen variation is relatively small, but as it gets larger, the spectra seem to fluctuate more around the mean.

\begin{figure*}[t]
\centering

\subfigure[]{
    \includegraphics[width=0.235\textwidth]{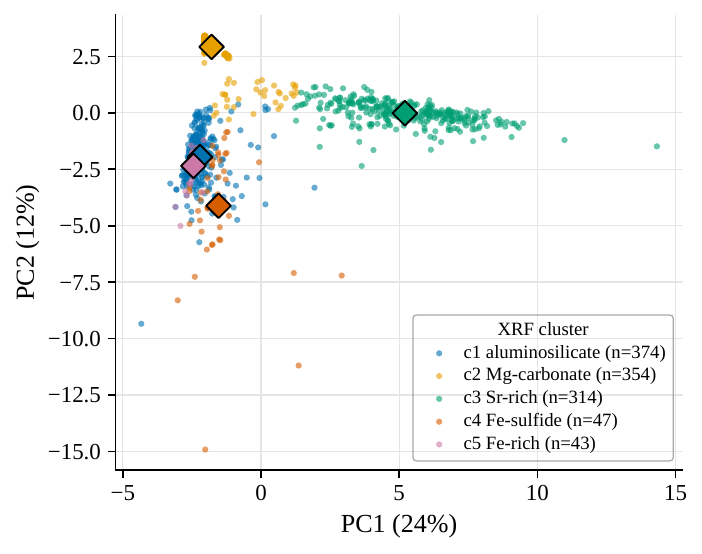}
    \label{fig:xrf_pca}
}
\hfill
\subfigure[]{
    \includegraphics[width=0.235\textwidth]{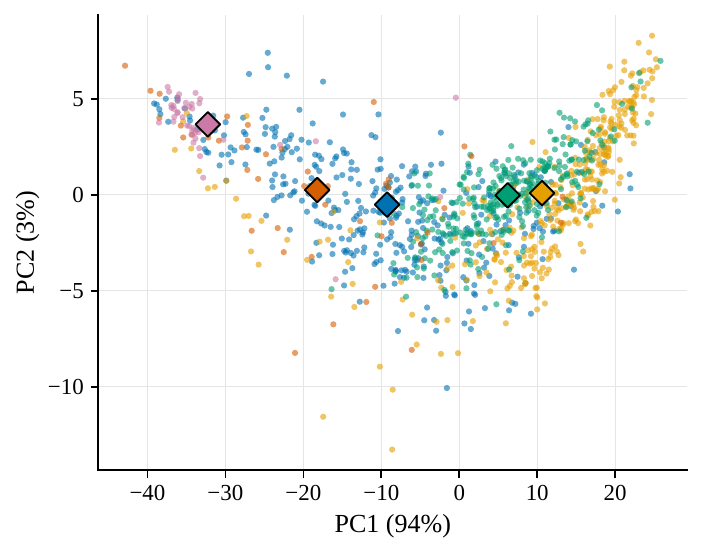}
    \label{fig:hsi_pca}
}
\hfill
\subfigure[]{
    \includegraphics[width=0.235\textwidth]{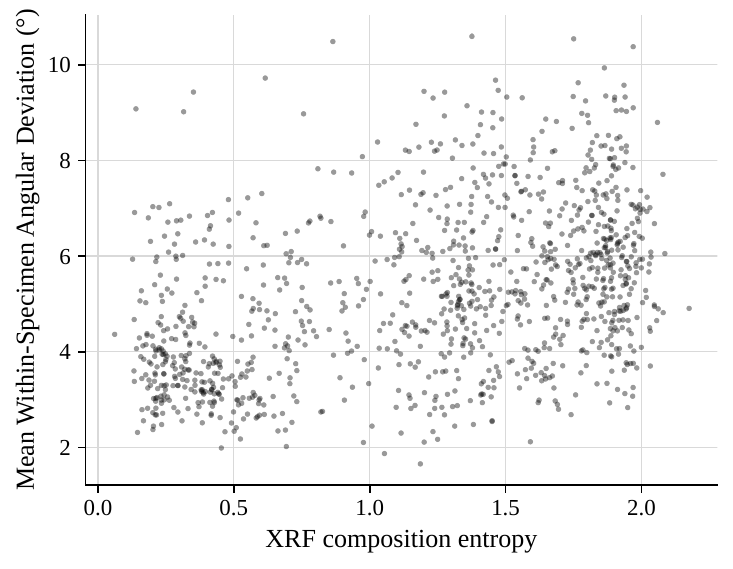}
    \label{fig:entropy_variance}
}
\hfill
\subfigure[]{
    \includegraphics[width=0.235\textwidth]{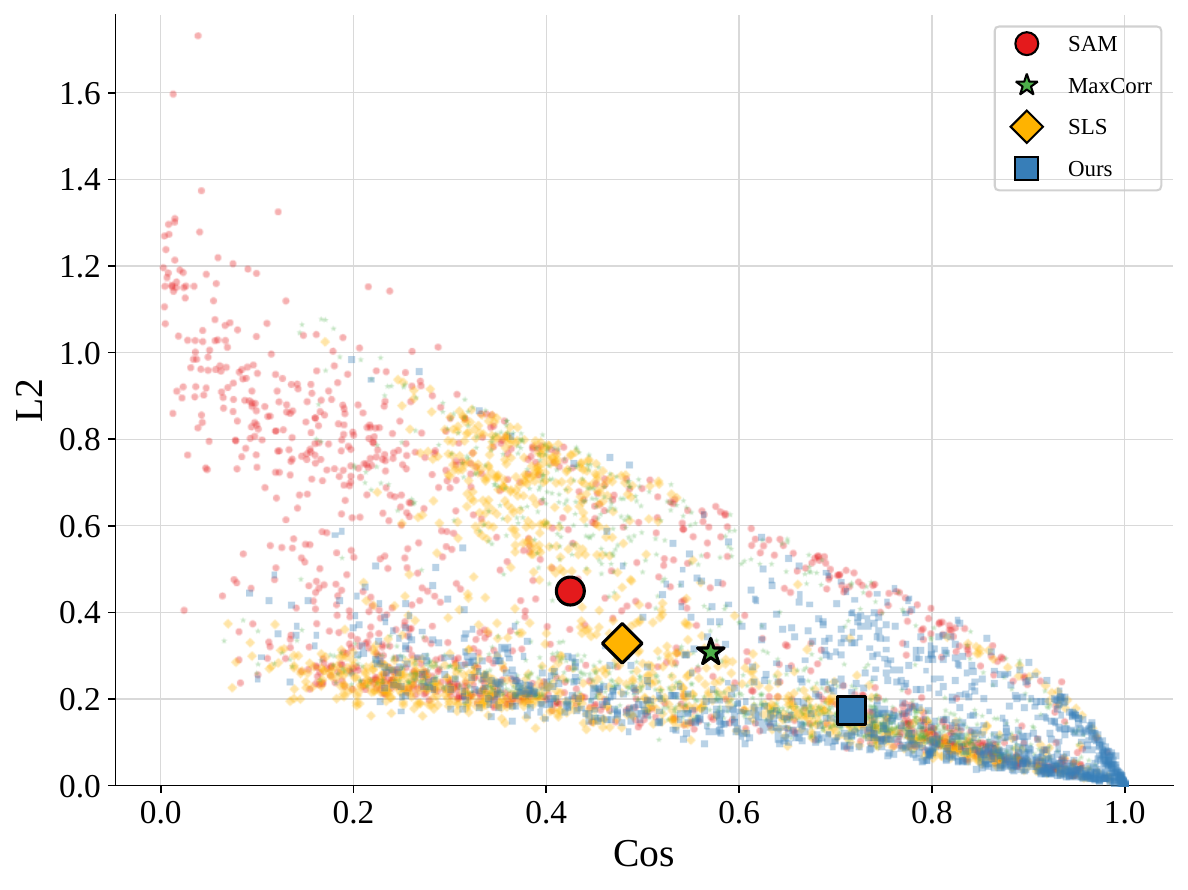}
    \label{fig:perf}
}

\vspace{-2.5mm}
\caption{\textbf{(a)} 
PCA projection of XRF composition vectors, grouped into five chemically coherent clusters.
\textbf{(b)} 
PCA projection of the mean hyperspectral spectrum of each specimen, coloured according to the XRF-derived clusters in (a)
(diamonds indicate cluster centroids of (a)). 
\textbf{(c)} MWAD and XRF composition entropy scatterplot. \textbf{(d)} Performance of our method and the baselines across all specimens, evaluated using the XRF measurements.}
\label{fig:separability}
\end{figure*}

\myitem{Cross-modal structure (HSI--XRF).} We investigate how both modalities behave jointly. In Fig.~\ref{fig:xrf_pca}, we scatterplot the $S$ XRF vectors in 2D, and cluster them in 5 groups (see 5 groups of colored dots). Keeping the color of each cluster, in Fig.~\ref{fig:hsi_pca} we scatterplot all the $\mathbf{X}_s$. We can see that there is in fact some underlying structure as $\mathbf{X}_s$ are somewhat similarly (not trivially though) clustered as their XRF vectors, suggesting that learning a function that differentiates them is an interesting avenue of research.
Moreover, in Fig.~\ref{fig:entropy_variance}, we investigate if there is correlation between: (i) ($x$-axis) the XRF vector entropy, defined as $H(\mathbf{y}_s)=-\sum_{j=1}^{N} y_{s,j}\log(y_{s,j})$, which quantifies whether the composition is concentrated around few elements (low $H$) or is more uniformly distributed across many (high $H$), and (ii) ($y$-axis) spectral variability, measured using
$\mathrm{MWAD}_s$ (defined in (\ref{eq:mwad}). We can observe that although there is a positive trend, it is not definitive; however, one can argue that when XRF is more ``random", pixels exhibit increased variability (upper right part of figure).


\section{Case Study: Characterizing Specimens from Hyperspectral Imagery}
\label{sec:case}
In this section, we define the composition estimation task on the \emph{Minerals in the Wild} dataset, along with two metrics to assess the performance of an algorithm on the task. We will propose a solution and compare it against simpler baselines.

\subsection{Task Formulation}

\myitem{Task.} We are interested in a function $f(\cdot)$ that receives as input
$\mathbf{X}_s\in\mathbb{R}^{H_s\times W\times B}$, and computes its elemental composition:
\begin{equation}
    f(\cdot): \mathbb{R}^{H\times W\times B} \rightarrow \mathbb{R}^{N},~\text{with}~f(\mathbf{X}_s)=\hat{\mathbf{y}}_s,
\end{equation}
where $\hat{\mathbf{y}}_s$ is the elemental composition of $s$ over $N$ elements.

\myitem{Metrics.} We evaluate the performance of $f(\cdot)$ by comparing the predicted composition $\hat{\mathbf{y}}_s$ against the XRF-derived ground truth $\mathbf{y}_s$ using the $\mathrm{L2}$ loss, which measures compositional error (accounting for direction and magnitude), and $\mathrm{Cos}$, which measures cosine similarity and hence agreement in direction.
\begin{empheq}[box=\fbox]{equation} \label{eq:metrics}
\begin{array}{c@{\hspace{0.8em}}c}
\mathrm{L2}(\hat{\mathbf{y}}_s,\mathbf{y}_s)
=
\left\|
\hat{\mathbf{y}}_s-\mathbf{y}_s
\right\|_2^2,
~ 
\mathrm{Cos}(\hat{\mathbf{y}}_s,\mathbf{y}_s)
=
\frac{\hat{\mathbf{y}}_s^{\top}\mathbf{y}_s}
{\|\hat{\mathbf{y}}_s\|_2\|\mathbf{y}_s\|_2}
\end{array}
\end{empheq}

\myitem{Preliminaries.} Throughout this section, matrix $\mathbf{U}\in\mathbb{R}^{L\times B}$ encodes the $L$ USGS mineral signatures~\cite{kokaly2017usgs}, resampled at the $B$ bands of our camera\footnote{The original USGS signatures are sampled at a finer spectral resolution than our camera, so we spectrally resample them to our camera bands, following the Gaussian spectral response function approach~\cite{yokoya2012multispectral}.}.
The input $\mathbf{X}$ and USGS library $\mathbf{U}$ are individually preprocessed using continuum removal~\cite{clark1984reflectance}. To simplify notation, we denote the resulting representations also as $\mathbf{X}$ and $\mathbf{U}$. 
In what follows, we drop subscript $s$.


\subsection{Proposed Pipeline}
\label{sec:method}

The pipeline consists of three stages, and is applied per specimen. \emph{First}, the specimen is decomposed into its pixels and, for each pixel, we estimate its composition under the USGS spectral basis, i.e., as a combination of mineral signatures from the USGS library. \emph{Second}, the pixel-level USGS compositions are spatially aggregated to get a specimen-level composition under the USGS basis. \emph{Third}, the specimen-level USGS composition is transformed to elemental composition\footnote{USGS contains compounds; we treat them exactly as the compounds appearing in the XRF exported vectors.}, enabling direct comparison with the XRF measurements.

\myitem{Pixel-level USGS compositions.} This is two procedures:

\noindent\textit{1) Candidates.} For each pixel $p\in\Omega$, we construct a reduced USGS library $\mathbf{U}_p=g(\mathbf{U},\mathbf{x}_p,t)$ by retaining signatures whose first- and second-order derivatives match those of $\mathbf{x}_p$.
Specifically, a library signature is selected if the cosine similarities between the corresponding derivatives exceed the threshold $t$ (excluding pairs with both negative similarities). The resulting library $\mathbf{U}_p\in\mathbb{R}^{L_p\times B}$ contains the $L_p$ candidate signatures used in the optimization.

\noindent\textit{2) Optimization.} The pixel-level USGS composition of specimen at pixel $p$ is represented by $\mathbf{z}_p^{\star}\in\mathbb{R}^{L_p}$, whose entries correspond to the weights assigned to the $L_p$ candidates; it is obtained by solving the convex optimization problem:
\begingroup
\setlength{\abovedisplayskip}{3pt}
\setlength{\belowdisplayskip}{3pt}
\begin{align}
\textbf{(OP)}\qquad
\mathbf{z}_{p}^{\star}
=
\arg\min_{\mathbf{z}\in\mathbb{R}^{L_p}}
&
\left\|
\mathbf{x}_p-\mathbf{z}^{\top}\mathbf{U}_p
\right\|_2^2
\label{eq:pixel_opt}
\\
\text{subject to}~~~
&
\mathbf{z}\ge\mathbf{0},
~
\mathbf{1}^{\top}\mathbf{z}=1.
\nonumber
\end{align}
\endgroup
Then, $\mathbf{z}_{p}^{\star}$ is embedded into the full USGS basis by assigning zero to all signatures excluded during filtering. By a slight abuse of notation, we denote this vector also by $\mathbf{z}_{p}^{\star}\in\mathbb{R}^{L}$.

\myitem{Specimen-level USGS composition.} 
We proceed to spatial averaging of $\mathbf{z}_{p}^{\star}$ across all pixels:
$\mathbf{z} =
\frac{1}{|\Omega|}
\sum_{p\in\Omega}
\mathbf{z}_{p}^{\star}$,
where $\mathbf{z}\in\mathbb{R}^L$ is a vector showing the weights of the specimen across the $L$ signatures, which sums to 1.

\myitem{Specimen-level elemental composition.} The $\mathbf{z}\in\mathbb{R}^{L}$ represents composition over the USGS basis, whereas the XRF ground truth $\mathbf{y}$ is over the elemental space, $\mathbb{R}^N$. 
So, we apply the harmonization procedure of Section~\ref{subsec:xrf} to convert USGS mineral composition into elemental. The resulting elemental vector is then projected onto the XRF elemental basis, discarding elements outside the basis and renormalizing, which yields the predicted elemental composition $\mathbf{\hat{y}}\in\mathbb{R}^N$.


\subsection{Experimental Evaluation}

We first check the effect of threshold $t$ to the performance of our algorithm and then compare it against simpler baselines.

\myitem{Effect of the filtering threshold $t$.} This parameter controls the strictness of the candidate-selection. For each value of $t$, we execute the complete pipeline (Section~\ref{sec:method}) and evaluate the resulting specimen-level compositions using the $\mathrm{L2}$ and $\mathrm{Cos}$ metrics, as defined in (\ref{eq:metrics}). We additionally report the \emph{coverage}, i.e., the percentage of specimens with at least one candidate USGS signature.
Table~\ref{tab:learn} shows the trade-off induced by $t$. Small values retain many candidates, increasing ambiguity during the optimization, whereas large values discard plausible signatures, eventually leaving specimens with no candidates. Consequently, coverage decreases with $t$. We use $t=0.10$ in the remainder, as it preserves full coverage while providing a balance between $\mathrm{Cos}$ and $\mathrm{L2}$.

\myitem{Performance comparison.} All baselines are instantiated within the proposed pipeline to enable specimen-level elemental prediction and validation against the XRF. They differ only in the pixel-level USGS composition estimation step. 
We used the CR versions of HSI and USGS spectra throughout. 
\begin{itemize}[noitemsep,topsep=0pt,leftmargin=*]
\item \textit{SAM.} Each pixel is assigned to the USGS signature with the minimum spectral angle.
\item \textit{MaxCorr.} Each pixel is assigned to the USGS signature with the maximum Pearson correlation.
\item \textit{SLS (simplex-constrained least squares).} For each pixel, we solve \textbf{(OP)} (\ref{eq:pixel_opt}) over the complete USGS, i.e., without the proposed candidate filtering.
\end{itemize}
To compare our method against the baselines, Fig.~\ref{fig:perf} visualizes the specimen-level performance in 2D plane, where a point corresponds to $(\text{$\mathrm{Cos}$,~$\mathrm{L2}$})_s^m$ for specimen $s$ and method $m \in \{\text{Ours},\text{SAM},\text{MaxCorr},\text{SLS}\}$. Hence, each method is represented by $S$ points, one per specimen. 
Ideally, we would like the specimens to concentrate in the lower right; and in fact, our algorithm tends to lie at that area more frequently compared to the other baselines. Moreover, we present Table~\ref{tab:comparison_all} which shows our method achieving the best performance across both $\mathrm{Cos}$ and $\mathrm{L2}$. Compared with \emph{MaxCorr}, our method improves the median $\mathrm{Cos}$ from $0.577$ to $0.773$ and reduces the median $\mathrm{L2}$ from $0.221$ to $0.150$. This suggests that restricting the optimization to pixel-dependent candidate signatures helps mitigate ambiguities introduced by the full library. 
Finally, in the last column, we present the average inference time needed per specimen; we can observe that our method is $6\times$ faster compared to the vanilla \emph{SLS} which optimizes using the whole USGS.

\begin{table}
\centering
\footnotesize
\setlength{\tabcolsep}{2.4pt}
\renewcommand{\arraystretch}{0.95}
\caption{Threshold selection of our method.}
\label{tab:learn}
\begin{tabular}{l ccc ccc c}
\toprule
& \multicolumn{3}{c}{$\mathrm{Cos}$ $\uparrow$}
& \multicolumn{3}{c}{$\mathrm{L2}$ $\downarrow$}
& \\
\cmidrule(lr){2-4}
\cmidrule(lr){5-7}
$t$ & Mean & Std & Med. & Mean & Std & Med. & Cov. \\
\midrule
0.05 & 0.659 & 0.201 & 0.680 & 0.202 & 0.148 & 0.155 & 100\% \\
\textbf{0.10} & \textbf{0.717} & 0.234 & 0.773 & \textbf{0.173} & 0.135 & 0.150 & \textbf{100\%} \\
0.15 & 0.711 & 0.270 & 0.809 & 0.215 & 0.220 & 0.166 & 97.1\% \\
0.20 & 0.746 & 0.257 & 0.838 & 0.214 & 0.223 & 0.168 & 84.7\% \\
0.30 & 0.767 & 0.255 & 0.861 & 0.226 & 0.223 & 0.163 & 61.9\% \\
0.40 & 0.749 & 0.264 & 0.853 & 0.276 & 0.255 & 0.181 & 36.7\% \\
0.50 & 0.774 & 0.244 & 0.885 & 0.278 & 0.264 & 0.177 & 17.0\% \\
0.60 & 0.844 & 0.209 & 0.929 & 0.205 & 0.193 & 0.155 & 6.6\% \\
\bottomrule
\end{tabular}
\end{table}

\begin{table}
\centering
\footnotesize
\setlength{\tabcolsep}{2.4pt}
\renewcommand{\arraystretch}{0.95}
\caption{Performance/accuracy and runtime, for all methods.}
\label{tab:comparison_all}
\begin{tabular}{l ccc ccc c}
\toprule
& \multicolumn{3}{c}{$\mathrm{Cos}$ $\uparrow$}
& \multicolumn{3}{c}{$\mathrm{L2}$ $\downarrow$}
& \multicolumn{1}{c}{Time $\downarrow$} \\
\cmidrule(lr){2-4}
\cmidrule(lr){5-7}
\cmidrule(lr){8-8}
Method & Mean & Std & Med. & Mean & Std & Med. & s/sample \\
\midrule
\emph{SAM}
& 0.425 & 0.278 & 0.344
& 0.449 & 0.316 & 0.348
& \textbf{3.83} \\

\emph{SLS}
& 0.479 & 0.214 & 0.421
& 0.329 & 0.232 & 0.236
& 33.710 \\

\emph{MaxCorr}
& 0.571 & 0.211 & 0.577
& 0.308 & 0.227 & 0.221
& 3.753 \\

\textbf{Ours}
& \textbf{0.717} & 0.234 & \textbf{0.773}
& \textbf{0.173} & 0.135 & \textbf{0.150}
& 5.521 \\
\bottomrule
\end{tabular}
\end{table}

\section{Conclusions}
\label{sec:conclusions}
We introduced a new dataset of naturally occurring mineral specimens, providing paired HSI and XRF measurements that enable the study of cross-modal relationships between spectral and elemental information. Furthermore, we developed a reproducible mineral identification pipeline, leveraging the HSI as input, the USGS library as support data, and the XRF as target. Finally, we proposed a convex optimization approach with a customized pruning filter and evaluated it against other USGS-informed benchmarks.

\bibliographystyle{IEEEbib}
\bibliography{references}

@misc{minerals-in-the-wild,
  title = {{Minerals in the Wild}},
  howpublished = {\url{https://github.com/EleftheriaTtl/minerals-in-the-wild}}
}

@techreport{iea2025energyai,
  author      = {{IEA}},
  title       = {{Energy and AI}},
  institution = {{IEA}},
  address     = {Paris, France},
  year        = {2025},
  url         = {https://www.iea.org/reports/energy-and-ai},
  note        = {Licence: CC BY 4.0}
}

@article{koirala2023multisensor,
  title={A multisensor hyperspectral benchmark dataset for unmixing of intimate mixtures},
  author={Koirala, Bikram and others},
  journal={IEEE Sensors Journal},
  year={2023}
}

@article{koirala2020robust,
  title={Robust supervised method for nonlinear spectral unmixing accounting for endmember variability},
  author={Koirala, Bikram and Zahiri, Zohreh and Lamberti, Alfredo and Scheunders, Paul},
  journal={IEEE TGRS},
  year={2020}
}

@article{rasti2022hapkecnn,
  title={HapkeCNN: Blind nonlinear unmixing for intimate mixtures using Hapke model and convolutional neural network},
  author={Rasti, Behnood and Koirala, Bikram and Scheunders, Paul},
  journal={IEEE TGRS},
  year={2022}
}

@article{heinz2001fully,
  title={Fully constrained least squares linear spectral mixture analysis method for material quantification in hyperspectral imagery},
  author={Heinz, Daniel C and others},
  journal={IEEE TGRS},
  year={2001}
}

@article{resmini1997mineral,
  title={Mineral mapping with hyperspectral digital imagery collection experiment (HYDICE) sensor data at Cuprite, Nevada, USA},
  author={Resmini, RG and others},
  journal={International Journal of Remote Sensing},
  year={1997},
  publisher={Taylor \& Francis}
}

@article{swayze2014mapping,
  title={Mapping advanced argillic alteration at Cuprite, Nevada, using imaging spectroscopy},
  author={Swayze, Gregg A and others},
  journal={Economic geology},
  year={2014}
}

@article{asadzadeh2025leveraging,
  title={Leveraging EnMAP hyperspectral data for mineral exploration: Examples from different deposit types},
  author={Asadzadeh, Saeid and Chabrillat, Sabine},
  journal={Ore Geology Reviews},
  pages={106912},
  year={2025},
  publisher={Elsevier}
}

@article{lessard2014oreSorting,
  author  = {Lessard, Joseph and de Bakker, Jan and McHugh, Larry},
  title   = {{Development of Ore Sorting and Its Impact on Mineral Processing Economics}},
  journal = {Minerals Engineering},
  year    = {2014},
  doi     = {10.1016/j.mineng.2014.05.019}
}

@misc{specim_swir,
  author       = {{Specim, Spectral Imaging Ltd.}},
  title        = {Specim SWIR Hyperspectral Camera},
  howpublished = {\url{https://www.specim.com/products/swir/}},
}

@book{goetz1985imaging,
  title={Imaging Spectrometry for Earth Remote Sensing},
  author={Goetz, Alexander F. H. and Vane, Gregg and Solomon, John E. and Rock, Barrett N.},
  year={1985},
  publisher={Science},
}

@misc{bruker_s1_titan,
  author       = {{Bruker}},
  title        = {{S1 TITAN Handheld XRF Analyzer for Elemental Analysis}},
  howpublished = {\url{https://www.bruker.com/content/dam/bruger-com/literature/S1-TITAN_Overview_Brochure.pdf}},
}

@article{kokaly2017usgs,
  title={USGS Spectral Library Version 7},
  author={Kokaly, Raymond F. and others},
  journal={U.S. Geological Survey Data Series},
  year={2017},
  publisher={U.S. Geological Survey},
  doi={10.3133/ds1035}
}

@article{iordache2011sparse,
  title={Sparse unmixing of hyperspectral data},
  author={Iordache, Marian-Daniel and Bioucas-Dias, Jos{\'e} M and Plaza, Antonio},
  journal={IEEE TGRS},
  year={2011}
}

@article{clark1984reflectance,
  author  = {Clark, Roger N. and Roush, Ted L.},
  title   = {Reflectance Spectroscopy: Quantitative Analysis Techniques for Remote Sensing Applications},
  journal = {Journal of Geophysical Research},
  volume  = {89},
  number  = {B7},
  pages   = {6329--6340},
  year    = {1984},
  doi     = {10.1029/JB089iB07p06329}
}

@article{yokoya2012multispectral,
  author  = {Naoto Yokoya and Akira Iwasaki and Kazuhiro Imai},
  title   = {Spectral Response Function Estimation for Hyperspectral Sensors},
  journal = {IEEE TGRS},
  year    = {2012}
}


\end{document}